\documentclass[runningheads]{llncs}
\usepackage{graphicx}
\usepackage{subfig}
\usepackage[cmex10]{amsmath}
\usepackage{hyperref}
\usepackage{cleveref}
\usepackage{rotating}
\usepackage{makecell}
\usepackage{float}
\makeatletter
\newif\if@blind
\@blindfalse 
\if@blind \def\hl#1{}\else
   \let\hl\relax
\fi

\makeatletter
\newcommand{\printfnsymbol}[1]{%
  \textsuperscript{\@fnsymbol{#1}}%
}
\makeatother

\begin{document}
\title{Few-shot Learning with Deep Triplet Networks for Brain Imaging Modality Recognition}
\titlerunning{Few-shot learning for brain imaging modality recognition}
%
\author{\hl{Santi Puch
\and Irina S\'anchez
\and Matt Rowe}}
\authorrunning{\hl{S. Puch et al.}}
%
\institute{\hl{QMENTA, Boston, MA, United States \\
\email{\{santi,irina,matt\}@qmenta.com}}}
\maketitle              
\begin{abstract}
Image modality recognition is essential for efficient imaging workflows in current clinical environments, where multiple imaging modalities are used to better comprehend complex diseases. Emerging biomarkers from novel, rare modalities are being developed to aid in such understanding, however the availability of these images is often limited. This scenario raises the necessity of recognising new imaging modalities without them being collected and annotated in large amounts. In this work, we present a few-shot learning model for limited training examples based on Deep Triplet Networks. We show that the proposed model is more accurate in distinguishing different modalities than a traditional Convolutional Neural Network classifier when limited samples are available. Furthermore, we evaluate the performance of both classifiers when presented with noisy samples and provide an initial inspection of how the proposed model can incorporate measures of uncertainty to be more robust against out-of-sample examples.

\end{abstract}
\section{Introduction}
In recent decades, many useful imaging biomarkers have emerged from multiple imaging modalities such as CT, PET, SPECT and MRI (and its many sub-modalities) to assist with differential diagnosis, disease monitoring and measuring the efficacy of pharmaceutical treatments. Diagnostic workflows and clinical trials have therefore become dependent on the simultaneous use of multiple modalities to augment the clinical understanding of complex diseases. This diversity of imaging modalities 
creates complexity for image archival systems such as PACS, VNAs and cloud-based solutions, and the institutions or businesses that use them. 

Classification of modalities and sub-modalities is important for efficient imaging workflows, a particularly difficult problem in MRI as the many distinct sub-modalities are not differentiated in a simple and consistent manner by image header information. 
For example, the use of contrast enhancing agents is a field often accidentally omitted or improperly populated in DICOM headers, meaning the use of a contrast enhancing agent can only be determined from the features of the image itself. In molecular imaging, an increasing variety of radioligands are being developed for monitoring different disease processes with each having distinct patterns of uptake or deposition. 
A human expert can easily distinguish them by their distinct visual features, however, scanner, vendor and center-specific idiosyncrasies in sequence implementation result in inconsistencies in DICOM header information that make automatic classification from DICOM headers alone highly challenging. 

Due to the importance of the visual features of the images to classify, the problem lends itself to Convolutional Neural Networks (CNNs), which have proved to be highly successful at achieving near human-level performance at classifying images based on visual features \cite{he2015delving}. 
A challenge to using CNNs for this kind of application is that they require large volumes of annotated data, which can be difficult to obtain for novel imaging biomarkers or rare modalities. For example, in a clinical trial utilising a novel imaging biomarker, it might be difficult to collect more than a handful of examples of the associated imaging sequence at startup. However, during the course of the trial, thousands of images may be acquired, requiring specific expertise to properly classify each sequence. 
Few-shot learning techniques offer a solution to creating robust classifiers from a limited amount of training data. 

In this paper, we propose a few-shot learning model based on Deep Triplet Networks, capable of capturing the most relevant imaging features that enable the differentiation between modalities even if the amount of training examples is limited.

\section{Methods}

\subsection{Data}
We collect a brain imaging dataset that consists of 7 MRI sequences (T1, T2, post-contrast T1, T2-FLAIR, PD, PASL and MRA), CT and FDG-PET imaging, sourced from several public datasets that include brain scans from healthy and diseased individuals. We consider two categories for these modalities: base modalities, that includes T1, T2, CT and FDG-PET, and are the most abundant and have the most distinctive imaging traits; and few-shot modalities, which includes T1-post, T2-FLAIR, PD, PASL and MRA modalities.

To train and evaluate the models, we extract 2D slices by sampling a normal distribution centered around the middle slice of the brain along the sagittal, coronal and axial axes. We sample 30874 slices of T1, 231759 of T2, 18541 of CT, 15432 of FDG-PET, 8017 of T1-post, 9828 of T2-FLAIR, 8370 of PD, 5321 of PASL and 8462 of MRA images. We used 70\% for training, 10\% for evaluation and 20\% for test.

\subsection{Deep Triplet Networks}
\label{sec:embeddings_cnn_triplet_loss}

We approach the few-shot learning problem with Triplet Networks \cite{metric_learning_triplet_networks}. A Triplet Network is a type of metric learning algorithm designed to learn a metric embedding $\phi(x)$ and a corresponding distance function $d(x, x')$ induced by a normed metric, so that given a triplet of samples $(x, x^+, x^-)$ and a similarity measure $r(x, x')$ that satisfies $r(x, x^+) > r(x, x^-)$, the learned distance function satisfies $d(x, x^+) < d(x, x^-)$. In essence, Triplet Networks learn to project samples in a embedding space in which similar samples are closer and dissimilar samples are farther apart with respect to a normed metric.

\begin{figure}[H]
    \caption{A Deep Triplet Network takes an \textit{anchor}, a \textit{positive} and a \textit{negative} sample, computes their embeddings with a deep CNN and then learns a distance function that satisfies the similarities between the samples of the triplet.}
    \centering
    \includegraphics[width=0.5\textwidth]{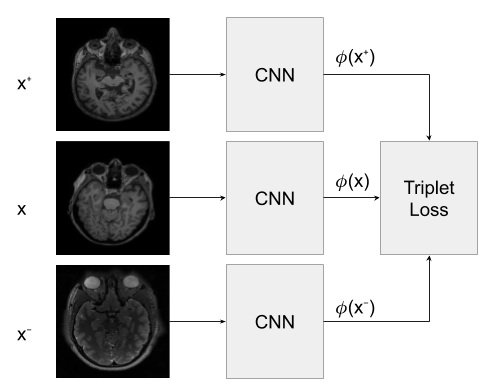}
    \label{fig:triplet_network_diagram}
\end{figure}

In our experimental setting, which corresponds to a multi-class image classification problem, the similarity measure $r(x, x')$ is defined by the labeling of our samples, that is, $r(x, x') = 1$ if $x$ and $x'$ belong to the same class and $r(x, x') = 0$ if $x$ and $x'$ belong to different classes. We define our distance function using the $L_1$ normed metric as follows:
\begin{equation}
    d(x, x') = ||\phi(x) - \phi(x')||_1
\end{equation}
where $\phi(x)$ is implemented with a deep CNN, hence the Deep Triplet Networks naming. Typically, the samples of the triplet $(x, x^+, x^-)$ are referred to as \textit{anchor}, \textit{positive} and \textit{negative}; the \textit{anchor} and \textit{positive} samples belong to the same class, while the \textit{negative} sample belongs to a different class. A diagram of a Deep Triplet Network is depicted in \autoref{fig:triplet_network_diagram}.

\subsection{Triplet loss with online hard-mining}

The loss used to train Deep Triplet Networks, referred to as triplet loss, is defined as follows:
\begin{equation}
    L(x, x^+, x^-) = max(d(x,x^+) - d(x, x^-) + m, 0) + \lambda (||x||_2 + ||x^+||_2 + ||x^-||_2)
\end{equation}
where $m$ is a margin that controls how much farther apart do we want the \textit{negative} sample to be with respect to the \textit{anchor} and \textit{positive} sample, and $\lambda$ is an hyperparameter that controls the amount of $L_2$ norm penalization of the embedding vectors.

We implement an online hard-mining triplet loss, which has been shown to be more efficient and help convergence \cite{online_triplet_loss}. Instead of computing the embeddings on the whole training set in an offline fashion and then mine the hard triplets, which satisfy $d(x, x^-) < d(x, x^+)$, we compute the embeddings on a mini-batch of $B$ images and then create a valid triplet with the hardest positive and the hardest negative for each anchor within that mini-batch \cite{online_triplet_loss_v2}. We choose a batch size of $B = 64$ as it provides a good balance between memory demand and a number of samples large enough to mine valid triplets among a variety of classes.

\subsection{Pipeline for image classification with Deep Triplet Networks}

\begin{figure}[H]
    \centering
        \caption{Diagram of the end-to-end pipeline for image classification with Deep Triplet Networks}
    \includegraphics[width=1\textwidth]{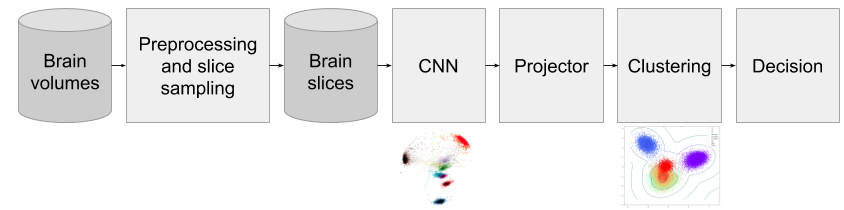}
    \label{fig:diagram_pipeline}
\end{figure}

We propose a pipeline for medical image volume classification based on Deep Triplet Networks. The pipeline, shown in \autoref{fig:diagram_pipeline}, starts with a preprocessing and slice sampling step that normalizes the orientation and image intensities of the volume and samples slices along the acquisition plane, emphasizing the sampling density around the FOV center. Each slice is then passed through a CNN that consists of a ResNet-50 \cite{residual_learning} initialized with pre-trained weights from ImageNet and trained with the triplet loss previously described. Then, the embedding vectors extracted per slice are projected to a lower-dimensional space using Principal Component Analysis (PCA), in order to remove the noisy components of the embedded representation \cite{dimensionality_reduction}. The PCA-projected embeddings are then clustered with a Gaussian Mixture Model (GMM) via expectation maximisation (EM). Unlike other clustering algorithms, such as k-means, a GMM is capable of capturing non-spherical cluster structures and provides estimates of the likelihood of a sample belonging to the model due to its probabilistic nature. We set the number of components of the GMM equal to the number of classes, and create a cluster to label mapping function by assigning to each cluster the most common class. From a GMM we can extract the posterior probability of each slice, that is, the probability that a sample came from each of the components of the mixture. We leverage that property to implement a hard decision function in which each slice is assigned the class with maximum probability, and the volume is classified by majority voting.

\section{Experiments}
\subsection{Hyperparameter search}
We use grid-search to obtain the optimal parameters of the model. The hyperparameters and options selected to optimize the network architecture are:
\begin{itemize}
\item Optimizer: ADAM, SGD with Nesterov momentum.
\item Learning Rate: $1e^{-3}$, $1e^{-4}$, $1e^{-5}$.
\item Learning Rate Decay: use a exponential decay with a decay rate of 0.9 every 1000 steps or not use decay.
\end{itemize}
The best performance was obtained when using SGD with Nesterov momentum as optimizer, a learning rate of $1e^{-3}$ and learning rate decay. We set in all experiments $L_2$ and $L_1$ regularization of weights to $1e^{-5}$ and $1e^{-6}$ respectively, the margin $m$ of the triplet loss to 2, the $L_2$ penalization of the embeddings $\lambda$ to 0.05, and the dimension of the embedding space to 64. We also perform random left-right and up-down flips as data augmentation.
This configuration is used to evaluate the performance of the proposed model in all the subsequent experiments.

Furthermore, for each set of experiments, we evaluate the PCA projector using different number of projection components in order to select the best configuration. After evaluating the results, we select PCA with 9 components. The GMM is configured so that each component has its own general covariance matrix.

\subsection{Few-shot learning}
We compare the performance of our proposed Triplet Network (TN) classifier against a standard CNN classifier when training with all the available data (exp1) and training with restrictions on the number of slices of the few-shot classes (exp2). In the latter, we restrict the number of slices of the few-shot classes to only 150 slices, which corresponds to 10 volumes from which 5 slices have been sampled for each of the 3 orthogonal axes. The CNN classifier is based on the same architecture and pre-trained weights than the TN classifier, plus a fully-connected layer to directly predict the class from the imaging data. 

\begin{figure}
    \centering
    \caption{Representation of the embedding space using the first three principal components of the evaluation embedding's projection on experiment 1 (left) and experiment 2 (right). Orange: T2, brown: T1, blue: CT, red: FDG-PET, purple: T2-FLAIR, yellow: T1-post, green: PD, pink: MRA, cyan: PASL.}
     \subfloat{
      \includegraphics[width=0.4\textwidth]{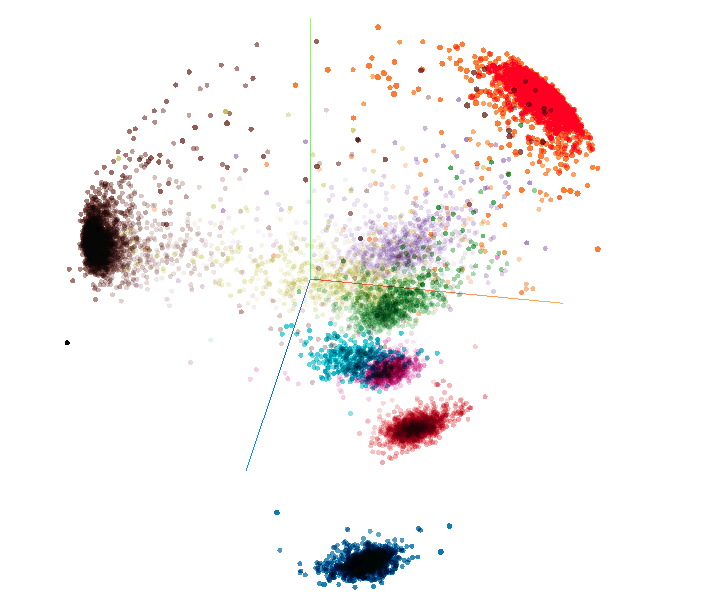}
     }
     \subfloat{
      \includegraphics[width=0.4\textwidth]{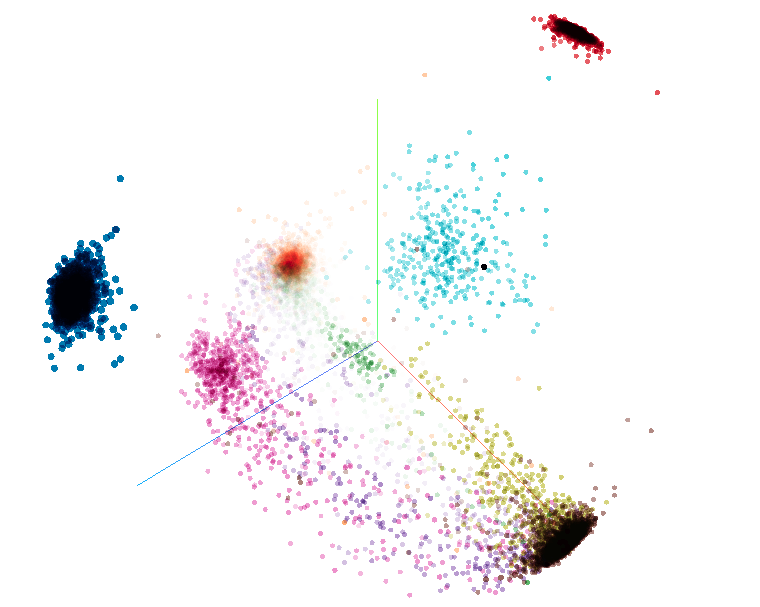}
     }
     \label{fig:embeddings}
  \end{figure}

In Table~\ref{tab:results} we present the class-wise average of the precision, recall and F1-score, and the balanced accuracy for both experiments. The standard CNN classifier performs considerably well when trained with all the available data, but is unable to capture the relevant imaging traits of the few-shot classes when the training data is scarce. However, the TN classifier is able to produce an embedding space (\autoref{fig:embeddings}) that separates the modalities into distinct clusters, allowing a better classification despite the under-representation of some classes. 

\begin{table}[]
\caption{Classification metrics of the Triplet Network classifier (TN) and the standard CNN classifier (CNN). B: base classes; F: few-shot classes.}
    \label{tab:results}
    \centering
    \resizebox{\textwidth}{!}{
    \begin{tabular}{|c|c|c|c|c|c|c|c|}
    \hline
    \textbf{Model}   & \textbf{Precision(B)} & \textbf{Recall(B)} & \textbf{F1-score(B)} & \textbf{Precision(F)} & \textbf{Recall(F)} & \textbf{F1-score(F)} & \textbf{Accuracy} \\ \hline
    \textbf{CNN classifier - exp 1} & 0.98                & 0.99             & 0.9875               & 0.966                 & 0.924              & 0.944                & 0.953              \\ \hline
    \textbf{TN classifier - exp 1}  & 1                & 0.938             & 0.965               & 0.89                  & 0.996              & 0.93                & \textbf{0.971}              \\ \hline
    \textbf{CNN classifier - exp 2} & 0.782                & 0.995               & 0.887                 & 1                 & 0.332              & 0.396                & 0.626              \\ \hline
   \textbf{TN classifier - exp 2}  & 0.92                & 0.967             & 0.942               & 0.816                 & 0.702              & 0.746                 & \textbf{0.819}              \\ \hline
    \end{tabular}
    }
\end{table}

\subsection{Robustness against noise}
We measure the robustness of both classifiers when the dataset is corrupted with additive gaussian noise and salt and pepper noise. We consider the scenario where the model has been trained with data that has been randomly corrupted by noise and tested with corrupted samples (exp3), and the scenario where the model has been trained with curated data but is also tested with corrupted samples (exp4). Further, we also analyze the performance when limiting the number of instances of the few-shot classes in both exp3 and exp4, as described in the previous section.

In Table~\ref{tab:results-noise} we show the class-wise average of the precision, recall and F1-score, and the balanced accuracy for the experiments where the data is corrupted with additive gaussian noise and salt and pepper noise. When the noise applied is additive gaussian, in both experiments and both scenarios (with and without limiting few-shot classes), the TN classifier outperfoms the CNN classifier, thus providing a more robust model. As expected, when the model has observed samples corrupted with noise during the training process the performance is better than when the training data is all curated. In the experiment using salt and pepper noise, when we use randomly corrupted samples during training the CNN classifier performs better than the TN classifier, but the results of the network decrease considerably when the few-shot classes are limited, while our proposed model is able to maintain a good performance. Both networks achieve bad results when are trained with curated data and tested with samples with salt and pepper noise. It is interesting to observe that the performance of the CNN classifier is similar with both types of noise, while the TN classifier has decreased substantially its performance when the noise used is salt and pepper.     


\begin{table}[]
\caption{Classification metrics of the few-shot classifier (Triplet) and the standard CNN classifier (Baseline) when trained with data corrupted noise (exp3) and trained with curated data but tested with corrupted volumes (exp4). B: base classes; F: few-shot classes.}
    \label{tab:results-noise}
    \centering
    \resizebox{\textwidth}{!}{
    \begin{tabular}{|c|c|c|c|c|c|c|c|}
    \hline
    \textbf{Model}   & \textbf{Precision(B)} & \textbf{Recall(B)} & \textbf{F1-score(B)} & \textbf{Precision(F)} & \textbf{Recall(F)} & \textbf{F1-score(F)} & \textbf{Accuracy} \\ \hline
    \textbf{Noise} & \multicolumn{7}{c|}{\textbf{Gaussian}} \\ \hline
    \textbf{CNN classifier - exp 3} & 0.99                & 0.987             & 0.987               & 0.956                 & 0.93              & 0.938                & 0.955              \\ \hline
    \textbf{TN classifier - exp 3}  & 0.992                & 0.947             & 0.97               & 0.888                  & 0.942              & 0.902                & \textbf{0.97}              \\ \hline
    \textbf{CNN classifier limit - exp 3} & 0.815                & 0.997               & 0.887                 & 1                 & 0.328              & 0.4                & 0.625              \\ \hline
   \textbf{TN classifier limit - exp 3}  & 0.942                & 0.965             & 0.952               & 0.658                 & 0.62              & 0.622                 & \textbf{0.773}              \\ \hline \hline
   \textbf{CNN classifier - exp 4} & 0.85                & 0.742             & 0.735               & 0.964                 & 0.478              & 0.638                & 0.596              \\ \hline
    \textbf{TN classifier - exp 4}  & 0.992                & 0.687             & 0.787               & 0.754                  & 0.774              & 0.682                & \textbf{0.737}              \\ \hline
    \textbf{CNN classifier limit - exp 4} & 0.725                & 10.817               & 0.732                 & 0.742                 & 0.194              & 0.29                & 0.47              \\ \hline
   \textbf{TN classifier limit - exp 4}  & 0.927                & 0.67             & 0.765               & 0.634                 & 0.678              & 0.588                 & \textbf{0.673}              \\ \hline
   \textbf{Model}   & \textbf{Precision(B)} & \textbf{Recall(B)} & \textbf{F1-score(B)} & \textbf{Precision(F)} & \textbf{Recall(F)} & \textbf{F1-score(F)} & \textbf{Accuracy} \\ \hline
   \textbf{Noise} & \multicolumn{7}{c|}{\textbf{Salt and pepper}} \\ \hline
    \textbf{CNN classifier - exp 3} & 0.982                & 0.985             & 0.982               & 0.946                 & 0.916              & 0.93                & \textbf{0.947}              \\ \hline
    \textbf{TN classifier - exp 3}  & 0.96                & 0.9375             & 0.945               & 0.658                  & 0.738              & 0.668                & 0.827              \\ \hline
    \textbf{CNN classifier limit - exp 3} & 0.765                & 0.99               & 0.87                 & 0.914                 & 0.31              & 0.384                & 0.625              \\ \hline
   \textbf{TN classifier limit - exp 3}  & 0.932                & 0.952             & 0.94               & 0.782                 & 0.75              & 0.756                 & \textbf{0.839}              \\ \hline \hline
   \textbf{CNN classifier - exp 4} & 0.832                & 0.612             & 0.647               & 0.822                 & 0.522              & 0.576                & \textbf{0.561}              \\ \hline
    \textbf{TN classifier - exp 4}  & 0.912                & 0.49             & 0.6325               & 0.798                  & 0.562              & 0.538                & 0.53              \\ \hline
    \textbf{CNN classifier limit - exp 4} & 0.785                & 0.505               & 0.602                 & 0.616                 & 0.26              & 0.25                & \textbf{0.47}              \\ \hline
   \textbf{TN classifier limit - exp 4}  & 0.722                & 0.49             & 0.545               & 0.64                 & 0.396              & 0.448                 & 0.44              \\ \hline
    \end{tabular}
    }
\end{table}

\subsection{Investigation of uncertainty measures}

We investigate the use of the estimated log-likelihood of a sample on the GMM model as a measure of uncertainty. To do so, we obtain a minimum log-likelihood threshold by taking the 1st percentile over the training data, which corresponds to a value of -12.44, and compare such threshold with the estimated log-likelihood of samples: a) that come from one of the classes on our dataset; b) that come from classes not represented in our dataset (e.g. volumes with binary masks or derived images). 

\begin{figure}
    \caption{Three samples of classes not represented in our dataset and a T1 slice, with their corresponding log-likelihood.}
    \centering
    \includegraphics[width=0.9\textwidth]{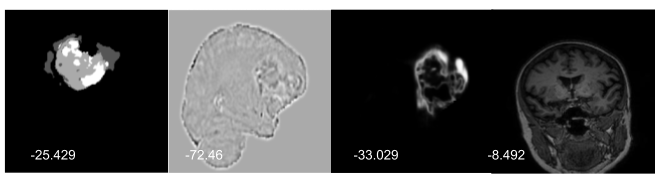}
    \label{fig:uncertainty}
\end{figure}

In \autoref{fig:uncertainty} we can see examples of the proposed experimental setting. We observe that a sample from a class represented in the dataset (in our case, a T1 volume from the test split) presents a log-likelihood value above the proposed threshold. However, samples from classes not represented in the dataset (concretely, a segmentation map, a filtered image and a probability map) have a log-likelihood value lower than the proposed threshold.

This basic observation serves as an initial validation of the possibility of having uncertainty estimates using the combination of a Deep Triplet Network and a GMM model, thus having the capability of discerning out-of-sample modalities.

\section{Conclusions}
We have provided evidence that Deep Triplet Networks are a viable solution for modality classification in a few-shot setting. The proposed model, when trained  with 30 times  less  instances  of  the  rarer  classes, surpasses substantially the performance of a CNN classifier trained under the same conditions. We have also concluded that the creation of an embedding space following a triplet network strategy increases the robustness against noise when compared to a standard CNN classifier. This is due to the fact that the results are not remarkably altered when the data is corrupted, regardless of whether the model has been trained with all the available samples or limiting the number of instances. Finally, we have explored the use of log-likelihood estimates of our model as a measure of uncertainty by evaluating such measure on samples not belonging to our dataset. We have found that this measure can effectively serve as an initial basis for uncertainty estimation, hence it can make our model more robust to unseen examples. This observation is preliminary and further investigation and development is required. Future work will focus on this topic, as well as extending the proposed model to alternative problems, such as disease staging.

\bibliographystyle{splncs04}
\bibliography{main}

\end{document}